%% file: gsaponaro-glu2017.tex
\pdfoutput=1 

\documentclass[a4paper]{article}

\usepackage{INTERSPEECH_v2}

\usepackage[T1]{fontenc} 
\usepackage[utf8]{inputenc} 
\usepackage[english]{babel} 

\usepackage[nolist]{acronym}
\usepackage{amsmath}
\usepackage{amssymb}
\usepackage{booktabs}
\usepackage{subfig}
\usepackage{tabularx}

\usepackage{tikz}
\usetikzlibrary{matrix,positioning} 
\usetikzlibrary{decorations.text}
\usetikzlibrary{shapes.geometric}
\tikzstyle{circle}=[shape=circle,minimum size=0.7cm,very thick]
\tikzstyle{every path}=[very thick]

\usepackage{hyperref}
\usepackage[update,prepend]{epstopdf}
\graphicspath{{figures/}}

\title{Interactive Robot Learning of Gestures, Language and Affordances}
\name{Giovanni~Saponaro$^1$, Lorenzo~Jamone$^{2,1}$, Alexandre~Bernardino$^1$, Giampiero~Salvi$^3$}
\address{
  $^1$Institute for Systems and Robotics\\Instituto Superior Técnico, Universidade de Lisboa, Lisbon, Portugal\\
  $^2$ARQ~(Advanced Robotics at Queen Mary)\\School of Electronic Engineering and Computer Science, Queen Mary University of London, UK\\
  $^3$KTH Royal Institute of Technology, Stockholm, Sweden}
\email{gsaponaro@isr.tecnico.ulisboa.pt, l.jamone@qmul.ac.uk, alex@isr.tecnico.ulisboa.pt, giampi@kth.se}

\newcommand{\AffWords}{Affordance--Words}
\newcommand{\eg}{e.\,g.}
\newcommand{\FB}{Forward--Backward}

\newcommand{\hh}{human--human}
\newcommand{\hr}{human--robot}
\newcommand{\hri}{\hr{} interaction}
\newcommand{\ie}{i.\,e.}
\newcommand{\ObjAct}{Object--Action}

\newcommand{\phmm}{\ensuremath{p_{\text{HMM}}}}
\newcommand{\pbn}{\ensuremath{p_{\text{BN}}}}

\begin{acronym}
\acro{AI}{Artificial Intelligence}
\acro{BN}{Bayesian Network}
\acro{HMM}{Hidden Markov Model}
\acro{MTRNN}{Multiple Timescales Recurrent Neural Network}
\acro{OAC}{\ObjAct{} Complex}
\acrodefplural{OAC}{\ObjAct{} Complexes}
\acro{PDF}{Probability Density Function}
\end{acronym}

\begin{document}

\maketitle
\begin{abstract} 
A growing field in robotics and \ac{AI} research is \hr{} collaboration, whose target is to enable effective teamwork between humans and robots. However, in many situations human teams are still superior to \hr{} teams, primarily because human teams can easily agree on a common goal with language, and the individual members observe each other effectively, leveraging their shared motor repertoire and sensorimotor resources. This paper shows that for cognitive robots it is possible, and indeed fruitful, to combine knowledge acquired from interacting with elements of the environment~(affordance exploration) with the probabilistic observation of another agent's actions.

We propose a model that unites (i)~learning robot affordances and word descriptions with (ii)~statistical recognition of human gestures with vision sensors. We discuss theoretical motivations, possible implementations, and we show initial results which highlight that, after having acquired knowledge of its surrounding environment, a humanoid robot can generalize this
knowledge to the case when it observes another agent~(human partner) performing the same motor actions previously executed during training.
\end{abstract}

\input{gsaponaro-glu2017-introduction}

\input{gsaponaro-glu2017-related_work}

\input{gsaponaro-glu2017-approach}

\input{gsaponaro-glu2017-results}

\input{gsaponaro-glu2017-conclusions}

\section{Acknowledgements}
This research was partly supported by the CHIST-ERA project IGLU and by the FCT project~UID/EEA/50009/2013.
We thank Konstantinos~Theofilis for his software and help permitting the acquisition of human hand coordinates in \hri{} scenarios with the iCub robot.

\bibliographystyle{IEEEtran}
\bibliography{glu2017_bibliography}

\end{document}

%% file: gsaponaro-glu2017-introduction.tex

\section{Introduction}

Robotics is progressing fast, with a steady and systematic shift from the industrial domain to domestic, public and leisure environments~\cite[ch.~65, Domestic Robotics]{siciliano:2016:handbook2}. Application areas that are particularly relevant and being researched by the scientific community include: robots for people's health and active aging, mobility, advanced manufacturing~(Industry~4.0). In short, all domains that require direct and effective \hri{} and communication (including language and gestures~\cite{matuszek:2014:aaai}).

However, robots have not reached the level of performance that would enable them to work with humans in routine activities in a flexible and adaptive way, for example in the presence of sensor noise, or unexpected events not previously seen during the training or learning phase. One of the reasons to explain this performance gap between \hh{} teamwork and a \hr{} teamwork is in the collaboration aspect, \ie, whether the members of a team understand one another. Humans have the ability of working successfully in groups. They can agree on common goals~(\eg, through verbal and non-verbal communication), work towards the execution of these goals in a coordinated way, and understand each other's physical actions~(\eg, body gestures) towards the realization of the final target. Human team coordination and mutual understanding is effective~\cite{ramnani:2004:natureneuro} because of~(i) the capacity to \emph{adapt} to unforeseen events in the environment, and re-plan one's actions in real time if necessary, and~(ii) a common motor repertoire and action model, which permits us to understand a partner's physical actions and manifested intentions as if they were our own~\cite{saponaro:2013:crhri}.

In neuroscience research, visuomotor neurons~(\ie, neurons that are activated by visual stimuli) have been a subject of ample study~\cite{rizzolatti:2001:nrn}. Mirror neurons are one class of such neurons that responds to action and object interaction, both when the agent acts and when it observes the same action performed by others, hence the name ``mirror''.

\begin{figure}
  \centering
  \includegraphics[width=0.9\columnwidth]{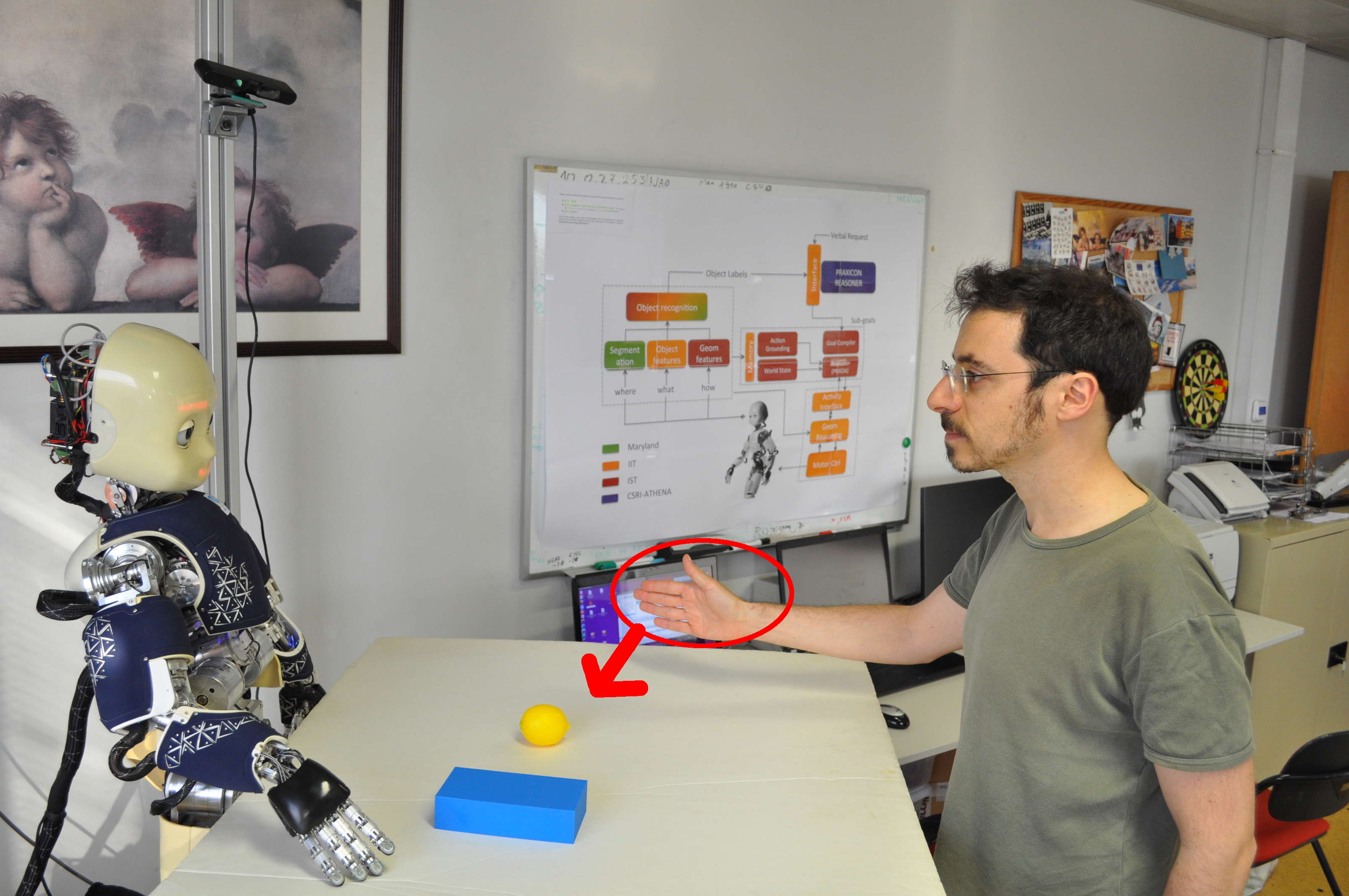}
  \caption{Experimental setup, consisting of an iCub humanoid robot and a human user performing a manipulation gesture on a shared table with different objects on top. The depth sensor in the top-left corner is used to extract human hand coordinates for gesture recognition. Depending on the gesture and on the target object, the resulting effect will differ.}
  \label{fig:experimental_setup}
\end{figure}

This work takes inspiration from the theory of mirror neurons, and contributes towards using it on humanoid and cognitive robots. We show that a robot can first acquire knowledge by sensing and self-exploring its surrounding environment~(\eg, by interacting with available objects and building up an affordance representation of the interactions and their outcomes) and, as a result, the robot is capable of generalizing its acquired knowledge while observing another agent~(\eg, a human person) who performs similar physical actions to the ones executed during prior robot training. Fig.~\ref{fig:experimental_setup} shows the experimental setup.

%% file: gsaponaro-glu2017-related_work.tex

\section{Related Work}

A large and growing body of research is directed towards having robots learn new cognitive skills, or improving their capabilities, by interacting autonomously with their surrounding environment. In particular, robots operating in an unstructured scenario may understand available opportunities conditioned on their body, perception and sensorimotor experiences: the intersection of these elements gives rise to object affordances~(action possibilities), as they are called in psychology~\cite{gibson:2014}. The usefulness of affordances in cognitive robotics is in the fact that they capture essential properties of environment objects in terms of the actions that a robot is able to perform with them~\cite{montesano:2008,jamone:2016:tcds}.
Some authors have suggested an alternative computational model called \acp{OAC}~\cite{kruger:2011:ras}, which links low-level sensorimotor knowledge with high-level symbolic reasoning hierarchically in autonomous robots.

In addition, several works have demonstrated how combining robot affordance learning with language grounding can provide cognitive robots with new and useful skills, such as learning the association of spoken words with sensorimotor experience~\cite{salvi:2012:smcb,morse:2016:cogsci} or sensorimotor representations~\cite{stramandinoli:2016:icdl}, learning tool use capabilities~\cite{goncalves:2014:icarsc,goncalves:2014:icdl}, and carrying out complex manipulation tasks expressed in natural language instructions which require planning and reasoning~\cite{antunes:2016:icra}.

In~\cite{salvi:2012:smcb}, a joint model is proposed to learn robot affordances~(\ie, relationships between actions, objects and resulting effects) together with word meanings. The data contains robot manipulation experiments, each of them associated with a number of alternative verbal descriptions uttered by two speakers for a total of 1270~recordings. That framework assumes that the robot action is known a~priori during the training phase~(\eg, the information ``grasping'' during a grasping experiment is given), and the resulting model can be used at testing to make inferences about the environment, including estimating the most likely action, based on evidence from other pieces of information.

Several neuroscience and psychology studies build upon the theory of mirror neurons which we brought up in the Introduction. These studies indicate that perceptual input can be linked with the human action system for predicting future outcomes of actions, \ie, the effect of actions, particularly when the person possesses concrete personal experience of the actions being observed in others~\cite{aglioti:2008:basketball,knoblich:2001:psychsci}. This has also been exploited under the deep learning paradigm~\cite{kim:2017:nn}, by using a \ac{MTRNN} to have an artificial simulated agent infer human intention from joint information about object affordances and human actions. One difference between this line of research and ours is that we use real, noisy data acquired from robots and sensors to test our models, rather than virtual simulations.

%% file: gsaponaro-glu2017-approach.tex

\section{Proposed Approach}

\begin{figure}
  \centering
  \includegraphics[width=0.8\columnwidth]{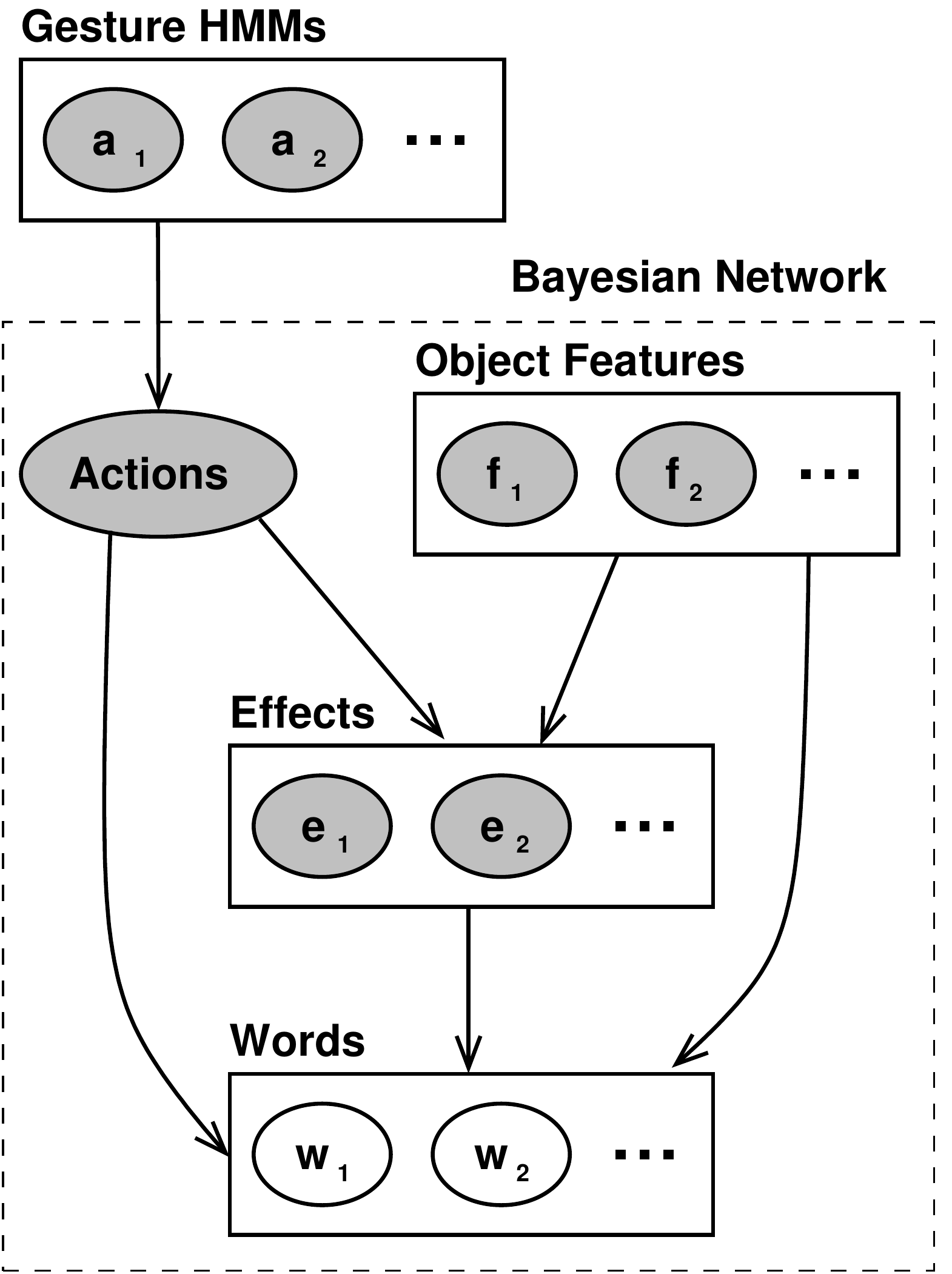}
  \caption{Abstract representation of the probabilistic dependencies in the model. Shaded nodes are observable or measurable in the present study, and edges indicate Bayesian dependency.}
  \label{fig:model}
\end{figure}

In this paper, we combine (1)~the robot affordance model of~\cite{salvi:2012:smcb}, which associates verbal descriptions to the physical interactions of an agent with the environment, with (2)~the gesture recognition system of~\cite{saponaro:2013:crhri}, which infers the type of action from human user movements.
We consider three \emph{manipulative gestures} corresponding to physical actions performed by agent(s) onto objects on a table~(see Fig.~\ref{fig:experimental_setup}): grasp, tap, and touch.
We reason on the effects of these actions onto the objects of the world, and on the co-occurring verbal description of the experiments. In the complete framework, we will use \acfp{BN}, which are a probabilistic model that represents random variables and conditional dependencies on a graph, such as in Fig.~\ref{fig:model}. One of the advantages of using \acp{BN} is that their expressive power allows the marginalization over any set of variables given any other set of variables.

Our main contribution is that of extending~\cite{salvi:2012:smcb} by relaxing the assumption that the action is known during the learning phase.
This assumption is acceptable when the robot learns through self-exploration and interaction with the environment, but must be relaxed if the robot needs to generalize the acquired knowledge through the observation of another~(human) agent.
We estimate the action performed by a human user during a \hr{} collaborative task, by employing statistical inference methods and \acp{HMM}. This provides two advantages. First, we can infer the executed action during training. Secondly, at testing time we can merge the action information obtained from gesture recognition with the information about affordances.

\subsection{\acl{BN} for \AffWords{} Modeling}
\label{sec:bn}

Following the method adopted in~\cite{salvi:2012:smcb}, we use a Bayesian probabilistic framework to allow a robot to ground the basic world behavior and verbal descriptions associated to it. The world behavior is defined by random variables describing: the actions~$A$, defined over the set~$\mathcal{A} = \{a_i\}$, object properties~$F$, over $\mathcal{F} = \{f_i\}$, and effects~$E$, over~$\mathcal{E} = \{e_i\}$. We denote~$X = \{A, F, E\}$ the state of the world as experienced by the robot. The verbal descriptions are denoted by the set of words~$W = \{w_i\}$. Consequently, the relationships between words and concepts are expressed by the joint probability distribution~$p(X,W)$ of actions, object features, effects, and words in the spoken utterance. The symbolic variables and their discrete values are listed in Table~\ref{tab:bnsymb}. In addition to the symbolic variables, the model also includes word variables, describing the probability of each word co-occurring in the verbal description associated to a robot experiment in the environment.

\begin{table}
    \centering
    \caption{The symbolic variables of the \acl{BN} which we use in this work~(a subset of the ones from~\cite{salvi:2012:smcb}), with the corresponding discrete values obtained from clustering during previous robot exploration of the environment.}
    \label{tab:bnsymb}
    \begin{tabular}{*{3}{l}} 
    \toprule
    name   & description     & values \\
    \midrule
    Action & action          & grasp, tap, touch \\
    Shape  & object shape    & sphere, box \\
    Size   & object size     & small, medium, big \\
    ObjVel & object velocity & slow, medium, fast \\
    \bottomrule
    \end{tabular}
\end{table}

This joint probability distribution, that is illustrated by the part of Fig.~\ref{fig:model} enclosed in the dashed box, is estimated by the robot in an ego-centric way through interaction with the environment, as in~\cite{salvi:2012:smcb}. As a consequence, during learning, the robot knows what action it is performing with certainty, and the variable~$A$ assumes a deterministic value. This assumption is relaxed in the present study, by extending the model to the observation of external~(human) agents as explained below.

\subsection{\aclp{HMM} for Gesture Recognition}

\newcommand{\myscalefactor}{0.8}

\newcommand{\standardhmm}[1]{
    \node[draw,circle] (hmm#1s1) {1};
    \node[draw,circle, right of=hmm#1s1] (hmm#1s2) {2};
    \node[circle, right of=hmm#1s2] (hmm#1s3) {\dots};
    \node[draw,circle, right of=hmm#1s3] (hmm#1s4) {Q};
    \node[left of=hmm#1s1]  (invisible1) {};
    \node[right of=hmm#1s4] (invisible2) {};
    \path[->] (hmm#1s1) edge (hmm#1s2);
    \path[loop above] (hmm#1s1) edge (hmm#1s1);
    \path[->] (hmm#1s2) edge (hmm#1s3);
    \path[loop above] (hmm#1s2) edge (hmm#1s2);
    \path[dashed] (hmm#1s2) -- (hmm#1s3);
    \path[->] (hmm#1s3) edge (hmm#1s4);
    \path[loop above] (hmm#1s4) edge (hmm#1s4);
    \path[->] (invisible1) edge (hmm#1s1);
    \path[->] (hmm#1s4) edge (invisible2);
}

\newcommand{\modeltwo}{
  \begin{tikzpicture}[scale=\myscalefactor, every node/.style={scale=\myscalefactor}]
  \matrix (M) [matrix of nodes, ampersand replacement=\&] {%
    grasp gesture HMM \& \standardhmm{1} \\
    tap gesture HMM \& \standardhmm{2} \\
    touch gesture HMM \& \standardhmm{3} \\
  };
  \end{tikzpicture}
}

\begin{figure}
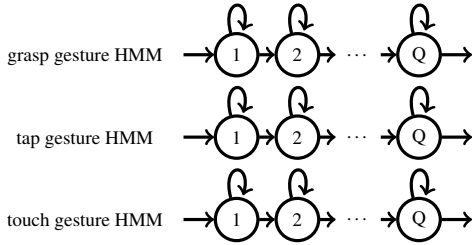

  \centering
  \modeltwo
  \caption{Structure of the \acp{HMM} used for human gesture recognition, adapted from~\cite{saponaro:2013:crhri}. In this work, we consider three independent, multiple-state \acp{HMM}, each of them trained to recognize one of the considered manipulation gestures.}
  \label{fig:hmms}
\end{figure}

As for the gesture recognition \acsp{HMM}, we use the models that we previously trained in~\cite{saponaro:2013:crhri} for spotting the manipulation-related gestures under consideration. Our input features are the 3D coordinates of the tracked human hand: the coordinates are obtained with a commodity depth sensor, then transformed to be centered on the person torso~(to be invariant to the distance of the user from the sensor) and normalized to account for variability in amplitude~(to be invariant to wide/emphatic vs narrow/subtle executions of the same gesture class).

The gesture recognition models are represented in Fig.~\ref{fig:hmms}, and correspond to the Gesture \acp{HMM} block in Fig.~\ref{fig:model}. The \ac{HMM} for one gesture is defined by a set of (hidden) discrete states~$\mathcal{S} = \{s_1, \dots, s_Q\}$ which model the temporal phases comprising the dynamic execution of the gesture, and by a set of parameters~$\lambda = \{ A, B, \Pi \}$, where~$A = \{ a_{ij} \}$ is the transition probability matrix, $a_{ij}$ is the transition probability from state~$s_i$ at time~$t$ to state~$s_j$ at time~$t+1$, $B = \{ f_i \}$ is the set of $Q$~observation probability functions~(one per state~$i$) with continuous mixtures of Gaussian values, and~$\Pi$ is the initial probability distribution for the states.

At recognition~(testing) time, we obtain likelihood scores of a new gesture being classified with the common \FB{} inference algorithm. In Sec.~\ref{sec:combination}, we discuss different ways in which the output information of the gesture recognizer can be combined with the \acl{BN} of words and affordances.

\subsection{Combining the \acs{BN} with Gesture \acsp{HMM}}
\label{sec:combination}

In this study we wish to generalize the model of~\cite{salvi:2012:smcb} by observing external~(human) agents, as shown in Fig.~\ref{fig:experimental_setup}. For this reason, the full model is now extended with a perception module capable of inferring the action of the agent from visual inputs. This corresponds to the Gesture \acp{HMM} block in Fig.~\ref{fig:model}. The \AffWords{} \acf{BN} model and the Gestures \acp{HMM} may be combined in different ways~\cite{pan:2006:ictai}:
\begin{enumerate}
\item the Gesture \acp{HMM} may provide a hard decision on the action performed by the human~(\ie, considering only the top result) to the \ac{BN},

\item the Gesture \acp{HMM} may provide a posterior distribution~(\ie, soft decision) to the \ac{BN},

\item if the task is to infer the action, the posterior from the Gesture \acp{HMM} and the one from the \ac{BN} may be combined as follows, assuming that they provide independent information:
\begin{equation*}
p(A) = \phmm(A) \, \pbn(A).
\end{equation*}
\end{enumerate}

%

%
%
%

In the experimental section, we will show that what the robot has learned subjectively or alone~(by self-exploration, knowing the action identity as a prior~\cite{salvi:2012:smcb}), can subsequently be used when observing a new agent~(human), provided that the actions can be estimated with Gesture \acp{HMM} as in~\cite{saponaro:2013:crhri}.

%% file: gsaponaro-glu2017-results.tex

\begin{figure}
    \centering
    \subfloat[][Prediction of the movement effect on a small sphere.]
    { \includegraphics[width=0.45\linewidth]{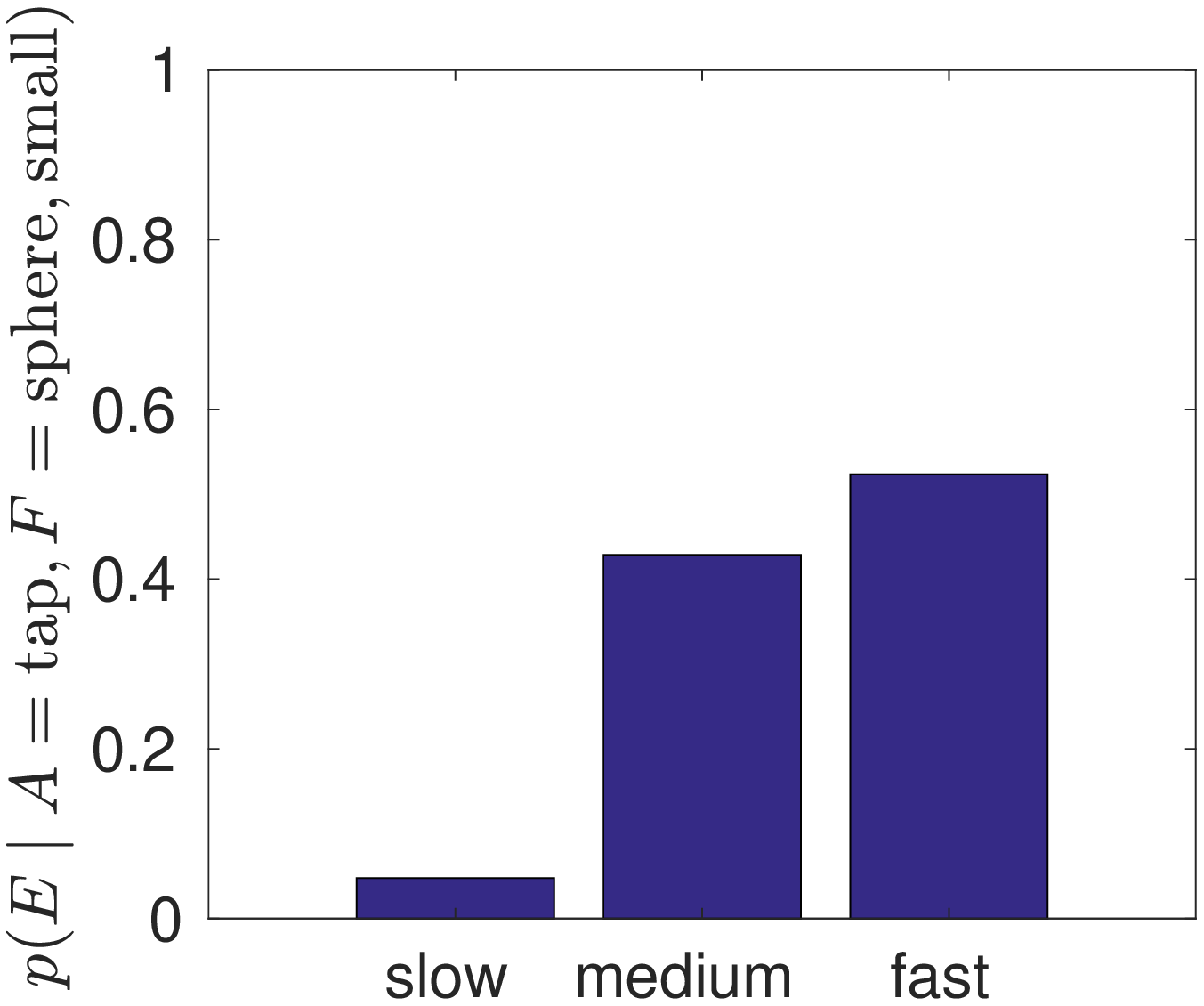} \label{fig:effect_pred:sphere} } \quad
    \subfloat[][Prediction of the movement effect on a big box.]
    { \includegraphics[width=0.45\linewidth]{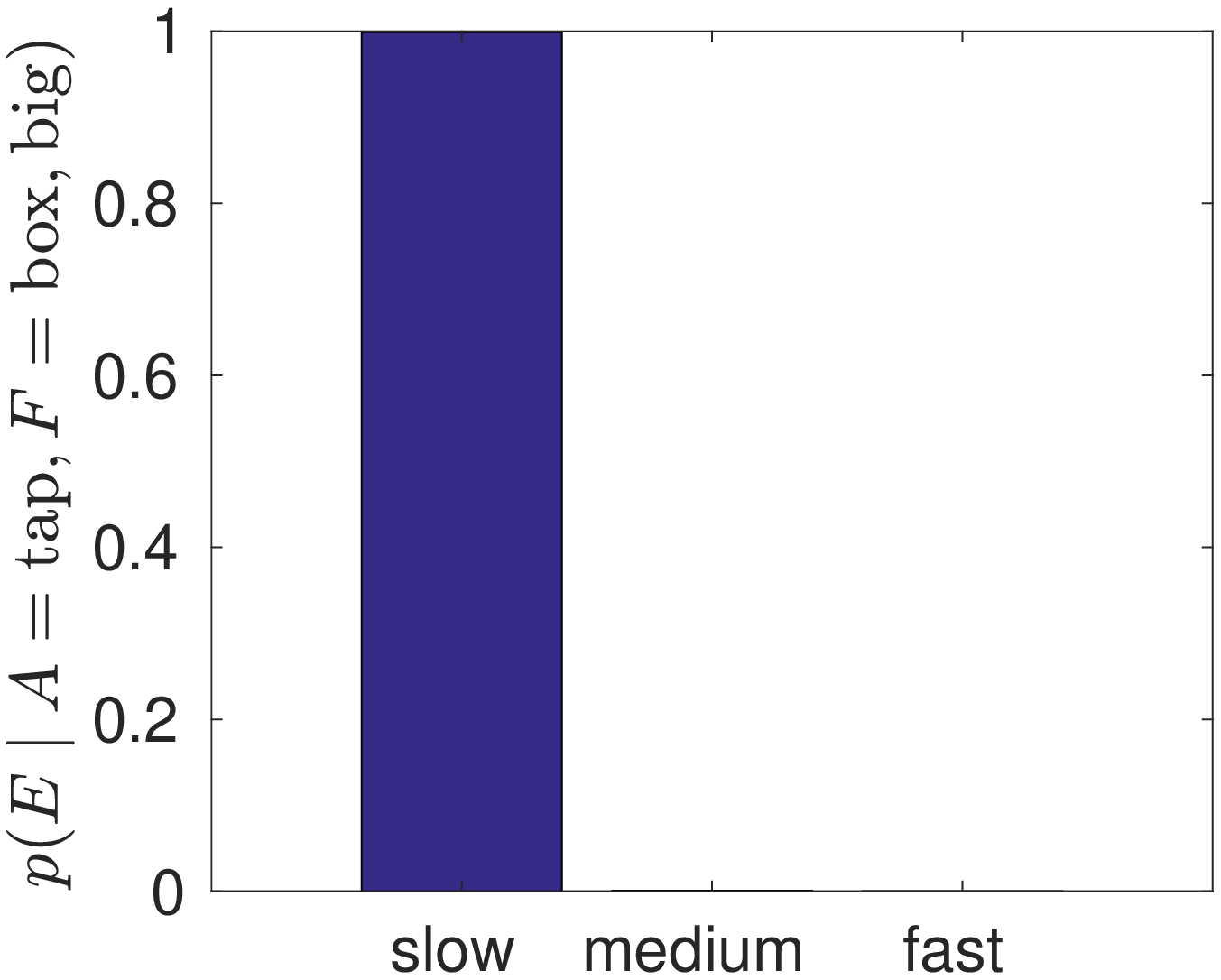} \label{fig:effect_pred:box} }
    \caption{Object velocity predictions, given prior information~(from Gesture \acp{HMM}) that the human user performs a tapping action.}
    \label{fig:effect_pred}
\end{figure}

\section{Experimental Results}

We present preliminary examples of two types of results: predictions over the effects of actions onto environment objects, and predictions over the associated word descriptions in the presence or absence of an action prior. In this section, we assume that the Gesture \acp{HMM} provide the discrete value of the recognized action performed by a human agent~(\ie, we enforce a hard decision over the observed action, referring to the possible combination strategies listed in Sec.~\ref{sec:combination}).

\subsection{Effect Prediction}

From our combined model of words, affordances and observed actions, we report the inferred posterior value of the Object Velocity effect, given prior information about the action~(provided by the Gesture \acp{HMM}) and also about object features~(Shape and Size). Fig.~\ref{fig:effect_pred} shows the computed predictions in two cases. Fig.~\ref{fig:effect_pred:sphere} shows the anticipated object velocity when the human user performs the tapping action onto a small spherical object, whereas Fig.~\ref{fig:effect_pred:box} displays it when the target object is a big box. Indeed, given the same observed action prior~(lateral tap on the object), the expected movement is very different depending on the physical properties of the target object.

\begin{figure}
\centering
\includegraphics[width=0.9\columnwidth]{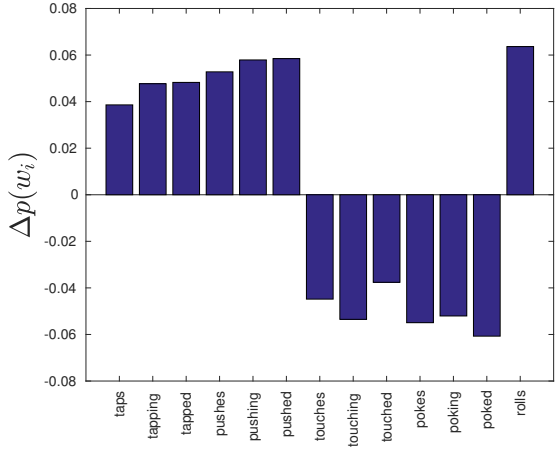}
\caption{Variation of word occurrence probabilities: $\Delta p(w_i) = p(w_i \mid F, E, A=\text{tap}) - p(w_i \mid F,E)$, where $F = \{\text{Size=big, Shape=sphere}\}$, $E = \{\text{ObjVel=fast}\}$. This variation corresponds to the difference of word probability when we add the tap action evidence~(obtained from the Gesture \acp{HMM}) to the initial evidence about object features and effects. We have omitted words for which no significant variation was observed.}
\label{fig:probdiff}
\end{figure}

\subsection{Prediction of Words}

In this experiment, we compare the associated \emph{verbal description} obtained by the \acl{BN} in the absence of an action prior, with the ones obtained in the presence of one. In particular, we compare the \emph{probability of word occurrence} in the following two situations:
\begin{enumerate}
\item when the robot prior knowledge~(evidence in the \ac{BN}) includes information about object features and effects only: \emph{Size=big, Shape=sphere, ObjVel=fast};

\item when the robot prior knowledge includes, in addition to the above, evidence about the action as observed from the Gestures \acp{HMM}: \emph{Action=tap}.
\end{enumerate}

Fig.~\ref{fig:probdiff} shows the variation in word occurrence probabilities between the two cases, where we have omitted words for which no significant variation was observed in this case. We can interpret the difference in the predictions as follows:
\begin{itemize}
\item as expected, the probabilities of words related to tapping and pushing increase when a tapping action evidence from the Gestures \acp{HMM} is introduced; conversely, the probabilities of other action words~(touching and poking) decreases;

\item interestingly, the probability of the word \emph{rolling}~(which is an effect of an action onto an object) also increases when the tapping action evidence is entered. Even though the initial evidence of case~$1$ already included some effect information~(the velocity of the object), it is only now, when the robot perceives that the physical action was a tap, that the event rolling is associated.
\end{itemize}

%% file: gsaponaro-glu2017-conclusions.tex

\section{Conclusions and Future Work}

Within the scope of cognitive robots that operate in unstructured environments, we have discussed a model that combines word affordance learning with body gesture recognition. We have proposed such an approach, based on the intuition that a robot can generalize its previously-acquired knowledge of the world~(objects, actions, effects, verbal descriptions) to the cases when it observes a human agent performing familiar actions in a shared \hr{} environment. We have shown promising preliminary results that indicate that a robot's ability to predict the future can benefit from incorporate the knowledge of a partner's action, facilitating scene interpretation and, as a result, teamwork.

In terms of future work, there are several avenues to explore. The main ones are (i)~the implementation of a fully probabilistic fusion between the affordance and the gesture components~(\eg, the soft decision discussed in Sec.~\ref{sec:combination}); (ii)~to run quantitative tests on larger corpora of \hr{} data; (iii)~to explicitly address the correspondence problem of actions between two agents operating on the same world objects~(\eg, a pulling action from the perspective of the human corresponds to a pushing action from the perspective of the robot, generating specular effects).